\documentclass{article}

\PassOptionsToPackage{numbers, compress}{natbib}

\usepackage[final]{neurips_2021}

\usepackage{natbib}

\usepackage[utf8]{inputenc} 
\usepackage[T1]{fontenc}    
\usepackage{hyperref}       
\usepackage{url}            
\usepackage{booktabs}       
\usepackage{amsfonts}       
\usepackage{nicefrac}       
\usepackage{microtype}      
\usepackage{xcolor}         

\title{Adversarial Conversational Shaping for Intelligent Agents}

\author{%
   Piotr Tarasiewicz \\
  University College London, UK\\
  \texttt{piotr.tarasiewicz.20@ucl.ac.uk} \\
   \And
   Sultan Kenjeyev\\
   University College London, UK \\
   \texttt{sultan.kenjeyev.20@ucl.ac.uk} \\
   \AND
   Ilana Sebag\\
   University College London, UK \\
   \texttt{ilana.sebag.20@ucl.ac.uk} \\
   \And
    Shehab Alshehabi\\
  University College London, UK\\
  \texttt{shehab.alshehabi.20@ucl.ac.uk} \\
}
\usepackage{booktabs}               

\usepackage{booktabs}               

\usepackage{natbib}
\usepackage{color, colortbl}

\usepackage{caption}

\usepackage{amsmath}
\usepackage{amsmath}
\usepackage{amssymb}
\usepackage{mathtools}

\usepackage{amssymb,graphicx,color}
\usepackage{amsfonts}
\usepackage{latexsym}
\usepackage{amsmath}

\usepackage{booktabs}               
\usepackage{amsfonts}               
\usepackage{nicefrac}               
\usepackage{microtype}              
\usepackage{soul}
\usepackage{xcolor}
\usepackage{mathtools}
\usepackage{geometry}
\usepackage{listings}
\usepackage{graphicx}
\usepackage{enumitem}

\usepackage{amsmath}
\usepackage{amssymb}
\usepackage[frozencache,cachedir=.]{minted}
\usepackage{textcomp}
\begin{document}

\maketitle


\begin{abstract}
The recent emergence of deep learning methods has enabled the research community to achieve state-of-the art results in several domains including natural language processing. However, the current robocall system remains unstable and inaccurate: text generator and chat-bots can be tedious and misunderstand human-like dialogue. In this work, we study the performance of two models able to enhance an intelligent conversational agent through adversarial conversational shaping: a generative adversarial network with policy gradient (GANPG) and a generative adversarial network with reward for every generation step (REGS) based on the REGS model presented in \citet{DBLP:journals/corr/LiMSRJ17}. This model is able to assign rewards to both partially and fully generated text sequences. We discuss performance with different training details : seq2seq \cite{sutskever2014sequence} and transformers \cite{vaswani2017attention} in a reinforcement learning framework. 

\end{abstract}

\section{Introduction}

Sequential data is ubiquitous: audio, video, text and time series are easily accessible to everyone. This allows the development of multiple state-of-the-art machine learning models in many fields and especially in natural language processing (NLP) \cite{GPT}. NLP has received intense interest in part due to the rapid rise of deep-learning-based methods. More specifically, nowadays NLP most used algorithms are based on recurrent neural networks (RNNs)\cite{RNNReport}, long short term memory (LSTM)\cite{LSTMPaper}, gated recurrent units (GRUs) \cite{GRUPaper}and bi-directional RNNs (BRNNs)\cite{BiDir}. Text generation lays the foundation for many applications such as open dialogue generation \cite{ritter-etal-2011-data, shen2018improving}, text summarizing \cite{DBLP:journals/corr/abs-2104-03057} and data augmentation \cite{feng2021survey} to name a few.

Most of the time, these systems are built upon an end-to-end model such as the sequence-to-sequence model (seq2seq) \cite{sutskever2014sequence} that aims to encode an input text sequence into a mathematical vector and then decoding the vector into a target text sequence. 

The objective of a dialogue system is to generate coherent and meaningful text responses given a dialogue input. The standard training method for such neural language models usually uses a maximum likelihood estimator (MLE) and an objective function derived from the Kullback-Leibler (KL) divergence between the empirical probability distribution representing the data and the parametric probability distribution output (\cite{labeau-cohen-2019-experimenting}). Despite the efficiency of this estimator, the conversational agent produces conventional answers that lack originality. Indeed, the MLE estimates the parameters of a probability distribution by maximizing the corresponding likelihood function so that under the assumed statistical model the observed data is more probable. Thus, it encourages the model to generates high-frequency words such as 'are', 'the', 'and', 'is' and it is harder to produce rare words and interesting answers.

In this work, we propose to compare existing works with a novel conversational agent that uses the T5 transformers, presented in \citet{raffel2020exploring}, during training. T5 is a pre-trained encoder-decoder model in which each task is converted into a human-language-like task. Also, to increase the creativity of our agent, we compare a generative adversarial network with policy gradient (GANPG) and a generative adversarial network with Reward for Every Generation Step (REGS) \cite{DBLP:journals/corr/LiMSRJ17} using pre-training only in a reinforcement learning framework.

\section{Related Work}

\textbf{Dialogue Generation} Open-domain dialogue generation is an increasingly prominent component of natural language processing (NLP). Indeed, nowadays, it constitutes the foundation of most NLP research. The techniques, used in NLP for text generation, have evolved alongside the progress in deep learning: variational auto-encoders are now widely used for text summarizing  \cite{miao-blunsom-2016-language} and dialogue modelling \cite{pmlr-v70-wen17a}. However, variational auto-encoder models have limitations due to posterior collapse (KL collapse). Multiple works, such as \cite{vinyals2015neural, serban2016generative, luan2016lstm} used the seq2seq method to build end-to-end conversational systems. Over the past years, GANs have also been one of the major NLP improvement  \cite{haidar2019textkdgan, rajeswar2017adversarial}. Finally, the combination of reinforcement learning (RL) and natural language processing is becoming omnipresent in the field \cite{ramamurthy2020nlpgym, luketina2019survey}. In the RL paradigm the agent's goal is to maximise the reward it receives from the environment. By designing the reward function in the environment we can optimise this setup for efficient word classification , accurate translation or dialogue generator.  \\

\noindent  \textbf{Generative Adversarial Networks} (GAN) are a Deep Learning technique that makes use of two neural networks : a generative model G and a discriminative model D that are trained simultaneously. G captures the distribution of the target data whilst D contributes to the training of G by classifying the generated data by G as real or machine-generated data. Adversarial Networks first appeared in 2014 \cite{goodfellow2014generative} as a pair of simple neural networks. This technique enables to generate new data with the same statistical properties as the input data used for the training set. Since then, Generative Adversarial Networks have been widely used in different fields especially in Computer Vision \cite{radford2016unsupervised}, \cite{NIPS2016_7c9d0b1f} and Natural Language Processing (NLP) \cite{glover2016modeling}, \cite{Li2017AdversarialLF}. GANs have also been combined with Reinforcement Learning framework in order to improve multiple generation tasks such as speech language generation with Policy Gradient Reinforcement Learning by back-propagating the error from the discriminator \cite{yu2017seqgan}. 

\noindent  \textbf{Transfer Learning} is widely used in the field of Machine Learning to reuse, transfer and leverage knowledge from a model. It is a popular approach in Deep learning where pre-trained models are used as the starting point on computer vision  \cite{ LI2020103853} and Natural Language Processing tasks \cite{ruder-etal-2019-transfer}. In our work, we use the transformers model T5 that was first presented by \cite{raffel2020exploring} in order to convert all text-based language problems into a text-to-text format.

\section{Generative Adversarial Networks (GANs) for Conversational Shaping}

Given a dialogue history $x$ , we aim to generate responses $y$ according to a policy defined by an encoder-decoder model. We defined two different models : a GAN with REGS and a GAN with Policy Gradient. Then, we will compare the usage of Seq2Seq and T5 Transformers on these models.

\subsection{GAN with Policy Gradient}

The GAN is composed of a generative model G and a discriminative model D. G defines a policy that generates the response $y$ by computing a probability vector of each token in the target sequence using a softmax function whilst D has the role of a classifier. In this case, we consider a binary discriminator that uses a sequence of pair of input and response dialogue $\{x,y\}$ and classify each input as either human generated or machine generated. Based on the work of \cite{li2015hierarchical}, we encode the input sequences into a vector of probabilities with the help of a Hierarchical Neural Auto-encoder. When the input is classify as human-generated, the corresponding assigned score is denoted by $Q_{+}(\{x,y\})$ and when the input is classify as machine-generated, the corresponding assigned score is denoted by $Q_{-}(\{x,y\})$. 

For the training, we use Policy Gradient algorithm : It is a Reinforcement Learning technique that aims at optimizing the parametrized policy with respect to the long-term cumulative reward by gradient descent. This training method encourages the model to generate human-like responses $y$ which are generated by sampling directly from the policy and used to update the discriminator.  In this framework, the assigned score is used as reward for the generator. Thus, the objective is to maximize the expected reward. We do so using the REINFORCE algorithm from \cite{Williams92simplestatistical} as in \cite{Li2017AdversarialLF}. $$ J(\theta) = E_{y \sim p(y \mid x)} ( Q_{+}(\{x,y\})\mid \theta) $$ And, its gradient can be estimated as follows :  $$ \nabla J(\theta) \approx [Q_{+}(\{x,y\}) - b(\{x,y\}) ] \nabla \log \pi (y \mid x) $$ $$ = [Q_{+}(\{x,y\}) - b(\{x,y\}) ] \nabla \sum_{t} \log p(y_{t} \mid x,y_{1:t-1}) $$

In the above equations, $\pi$ represents the probability of the generated responses $y$ and $b$ is the baseline used to regulate the variance of the estimate to make it efficient (low variance, unbiased and consistent). 

\subsection{GAN with Reward for Every Generation Step (REGS)}

In the previous section, we saw that the Generative Adversarial Network (GAN) with Policy Gradient model presents some flaws : a unique reward is assigned to each token of the human-generated response whilst we would expect different rewards. Also, in REGS, the discriminative model aims at assigning rewards to both fully and partially generated text sequences whilst the neural network uses the mean squared loss between the machine-generated text and real text rewards. This proves the importance of computing the reward at each intermediary step of the generation. \cite{DBLP:journals/corr/LiMSRJ17} propose two strategies to compute such rewards : Monte Carlo Search and training a discriminator that is able to assign rewards to partially and fully generated sequences. In this section, we focus on reproducing the latter strategy from \cite{DBLP:journals/corr/LiMSRJ17}'s paper.

Like in \cite{rajeswar2017adversarial}, we denote the generated sequences $\{y_{1:t}\}^{N_{Y}}_{t=1}$ and separate them into positive and negative sequences to denote the partially generated sequences, namely $\{y_{1:t}^{+}\}^{N_{Y+}}_{t=1}$  and $\{y_{1:t}^{-}\}^{N_{Y-}}_{t=1}$. Then, we randomly sample one example from each sequence and use it to update the discriminator.   For each partially generated sequence $Y_{t}$, the discriminator assigns a score  $Q_{+}(x,Y_{t})$ that will classify the sequence. The corresponding baseline value denoted $b(x,Y_{t})$ helps regulate the variance \cite{ranzato2016sequence}.

Hence, the generator is updated according to : 
$$ \nabla J(\theta) \approx \sum_{t}( Q_{+}(x,Y_{t}) - b(x,Y_{t})) \:\:\:\:\:\:\:\:\:\:\:\:\:\:\:\: \:\:\:\:$$ $$\:\:\:\:\:\:\:\:\ \:\:\:\:\:\:\:\: \nabla \log  p(y_{t} \mid x, Y_{1:t-1})$$

\subsection{Training details}

Given the dialogue history, we start by pre-training the generator by predicting target sequences. Then we train both Seq2Seq with attention and transformers T5 models for each neural network. 

We experimented several procedures in order to assess efficiency of the models : 
\begin{itemize}
    \item We trained the networks with and without teacher forcing.
    \item We trained the networks with and without layer freezing. When we applied layer freezing, we froze all layers of the model except the language model head and last decoder block of the transformer. 
    \item We tried different range of learning rates : $1e^{-3}$, $3e^{-4}$ and $ 5e^{-5}$. After several trials, we observed that the most suited learning rate was $1e^{-3}$. 
\end{itemize}

\subsubsection{Sequence-to-sequence (Seq2Seq) with attention}

In this section, we evaluate the pretrained and REGS models with Seq2seq with attention. We used the code from \cite{Li2017AdversarialLF}. The Seq2Seq framework relies on the encoder-decoder paradigm : it consists on encoding the source sequences and decoding the target sequence. The attention mechanism enforces the model to pay more attention on specific parts of the source sequence when decoding. That is, the encoder does not have to encode the entire input sequence into a vector and we are not relying only on the hidden vector anymore \cite{bahdanau2016neural}.

During training, at each time step, the decoder will generate a probabilistic vector $p_{i} \in \mathbb{R}$ that contains the probabilities of each token at each relevant time step. Then, given the input sequence $x_{i}$, we can compute the probability of some target sequence $y_{i}$ : $\mathbb{P}(y_{1}, ..., y_{m}) = \Pi_{i=1}^{m} p_{i}[y_{i}]$;  where $p_{i}[y_{i}]$ means that we extract the $y_{i}$th entry of the vector $p_{i}$ from the $i$th decoding step. We aim at generating human-like text, that is, maximizing the probability of the target sequence. This is the same as minimizing the standard cross entropy between the target distribution and the actual output : $$ - \log \mathbb{P}(y_{1}, ..., y_{m}) = - \log \Pi_{i=1}^{m} p_{i}[y_{i}]$$$$ = - \sum_{i=1}^{m} \log p_{i}[y_{i}]$$

\subsubsection{Transformers T5}

We also built our neural networks using a Transformers T5 for comparison purpose. T5 is an encoder-decoder model was first presented in \cite{raffel2020exploring}'s work. It is pre-trained on a multi-task mixture of unsupervised and supervised tasks and treat every text processing problem as a “text-to-text” problem. The advantage of the text-to-text format is that we can apply the same model, loss function, hyper-parameters, training procedure and decoding procedure on both the input and the output. 

\begin{figure}[h!]
	\centering
    \includegraphics[width=8cm]{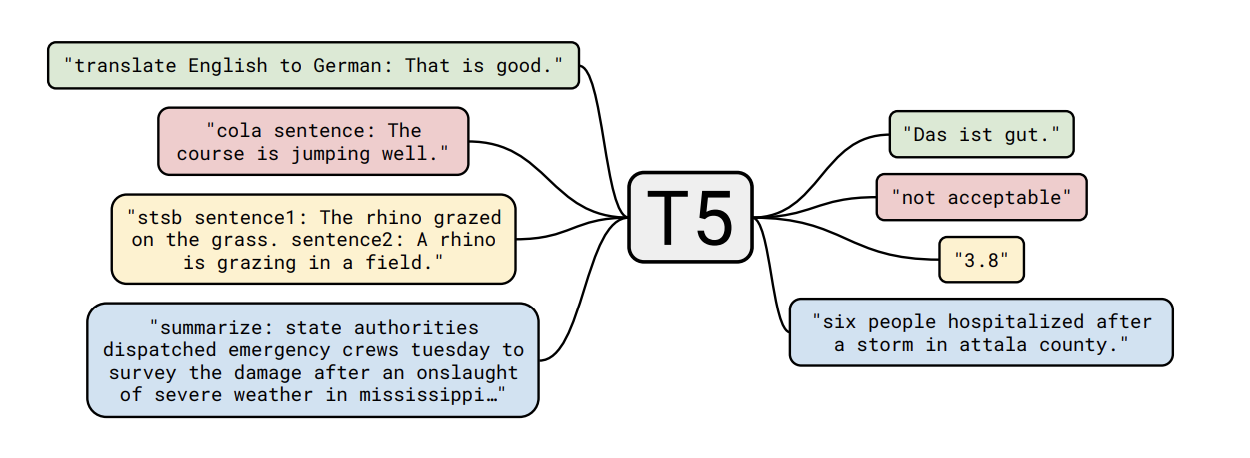} 
    \caption{Explanatory diagram of the T5 framework : “Text-to-Text Transfer Transformer” , image from \cite{raffel2020exploring} }
\end{figure}

We used the language model for both the generator and the discriminator. Indeed, as suggested in \cite{raffel2020exploring}, using the language model instead of a binary classifier would improve the accuracy of the predictions in this case. 
 
\subsubsection{Teacher Forcing}

Teacher Forcing is widely used in the field of Deep Learning Language Models to quickly and efficiently train RNNs that use the ground truth from a prior time step as input. This method consists on supplying observed sequence values as inputs during training and using the network’s own one-step ahead predictions to do multi-step sampling as explained in \cite{lamb2016professor}. For instance, we would give both human and machine generated responses to the generator for model updates and arbitrarily assign the value of 1 a s reward to each human generated response. That is, the 'teacher' is forcing the model to learn what is a human-like generated response. The idea of this method is to enforce the model the regularize itself when it deviates from the training dataset. 

In our experiments, we tried to implement our models with and without Teacher Forcing. It presents some advantages and disadvantages. This method will enforce a faster converging training as the hidden states of the model are not updated with wrong prediction sequences anymore, nonetheless, the issue of Exposure Bias occurs : \cite{Schmidt19}. When no ground-truth is available, there is a train-test discrepancy that might lead to inefficiency of the model.

\section{Experimental Results and Discussions}

We evaluate the above detailed methods on the Daily Dialogue dataset \cite{li2017dailydialog} and asses their efficiency using adversarial evaluation. We trained the T5 discriminator from scratch and raised the accuracy of predicting real and fake generated responses on balanced data. We also did the same procedure manually by denoting ourselves whether the dialogue and the corresponding generated answer made sense or not. We obtained the following results :

\begin{center}
\begin{tabular}{cccc}
\toprule 
 & T5 PT & T5 REGS &T5 PG\\
\midrule
\rowcolor {black!20} Dist-1	&0.2&	0.085&	0.079 \\
Dist-2&	0.47&	0.177&	0.162 \\
\rowcolor {black!20}Bleu-1 (1e-3)&	98.88&	58.26&56.84\\
Bleu-2	&55.2&	26.66&	24.96 \\
\rowcolor {black!20} Bleu-3	&21.01&	8.57&	6.84\\
Bleu-4&	14.21&	6.1	&1.07 \\
\rowcolor {black!20} Adversarial Accuracy&	68\%&	70\%&	72\%\\
Human Accuracy(avg)&	64\%&	73\%&	68\%\\
\bottomrule
\end{tabular} 
\captionof{table} {Final Results for T5 (PG stands for Policy Gradient and PT stands for pretrained)} 
\end{center}

\begin{center}
\begin{tabular}{ccc}
\toprule 
 & S2S PT &S2S REGS\\
\midrule
\rowcolor {black!20} Dist-1	&0.078	&0.081 \\
Dist-2&	0.551&	0.502 \\
\rowcolor {black!20}Bleu-1 (1e-3)&	30.06&	41.74\\
Bleu-2	&15.8&	23.3 \\
\rowcolor {black!20} Bleu-3	&10.16&	4.22\\
Bleu-4&	5.11&1.32 \\
\rowcolor {black!20} Adversarial Accuracy& 85\%&	75\%\\
Human Accuracy(avg)&	81\%&	80\%\\
\bottomrule
\end{tabular} 
\captionof{table} {Final Results for Seq2Seq (PT stands for pretrained)} 
\end{center}

\section{Conclusion}

In this work, we draw intuitions from the work of \cite{Li2017AdversarialLF}, \cite{xu-etal-2018-diversity} and \cite{DBLP:journals/corr/abs-2004-14507}. We propose an adversarial training approach for response generation. We built two Generative Adversarial Networks using the brand new text-to-text transformers T5 and compared it with a Seq2Seq implementation and a pretrained model. For the evaluation, we used BLUE and DIST. We choose cast the models in a Reinforcement Learning framework and deal with assigned scores as rewards for the classifiers. This allowed the generator to generate human-like dialogue responses. 

From the Tables 1 and 2 displayed above, we can conclude that out of all the presented models, the pretrained T5 model is the one that performs the best with the lowest adversarial accuracy of 68\%. Furthermore, we notice that all T5-based networks perform better than the Seq2Seq-based networks with a lowest adversarial accuracy of 80\% obtained with the Seq2Seq REGS model. We can assume that T5 performs better Seq2Seq as it is a powerful model, thus, the discriminator might easily cheat by finding hacks. Finally, we can observe that the GAN seems to contribute more to the efficiency of the model when paired with seq2seq rather than with T5.   

We also manually assessed the models by acting like the discriminator ourselves and classifying the generated response as real or fake given an input dialogue. We obtained the same ranking : pretrained T5 seems to perform the best and all T5-based models perform better than Seq2Seq-based models. 

\subsection{Future work}

We identified two main axes of expansions for this project. On the first hand, implementing a Diversity-Promoting GAN would allows to assess and compare the efficiencies of the models. On the other hand, exploring and applying counterfactual reasoning to these models could show great improvement. 

\textbf{Diversity-Promoting Generative Adversarial Network (DP-GAN)} : \cite{xu-etal-2018-diversity} implemented a DP-GAN in a Reinforcement Learning framework. This model is contains a language model based discriminator D trained over real and generated text, in opposition to our binary classifier, and assigns low reward to repetitive text and high reward for novel and rare text, to encourage the generator to produce more diverse text.The output of the discriminator is used as reward for the generator.

\begin{figure}[h!]
	\centering
    \includegraphics[width=7cm]{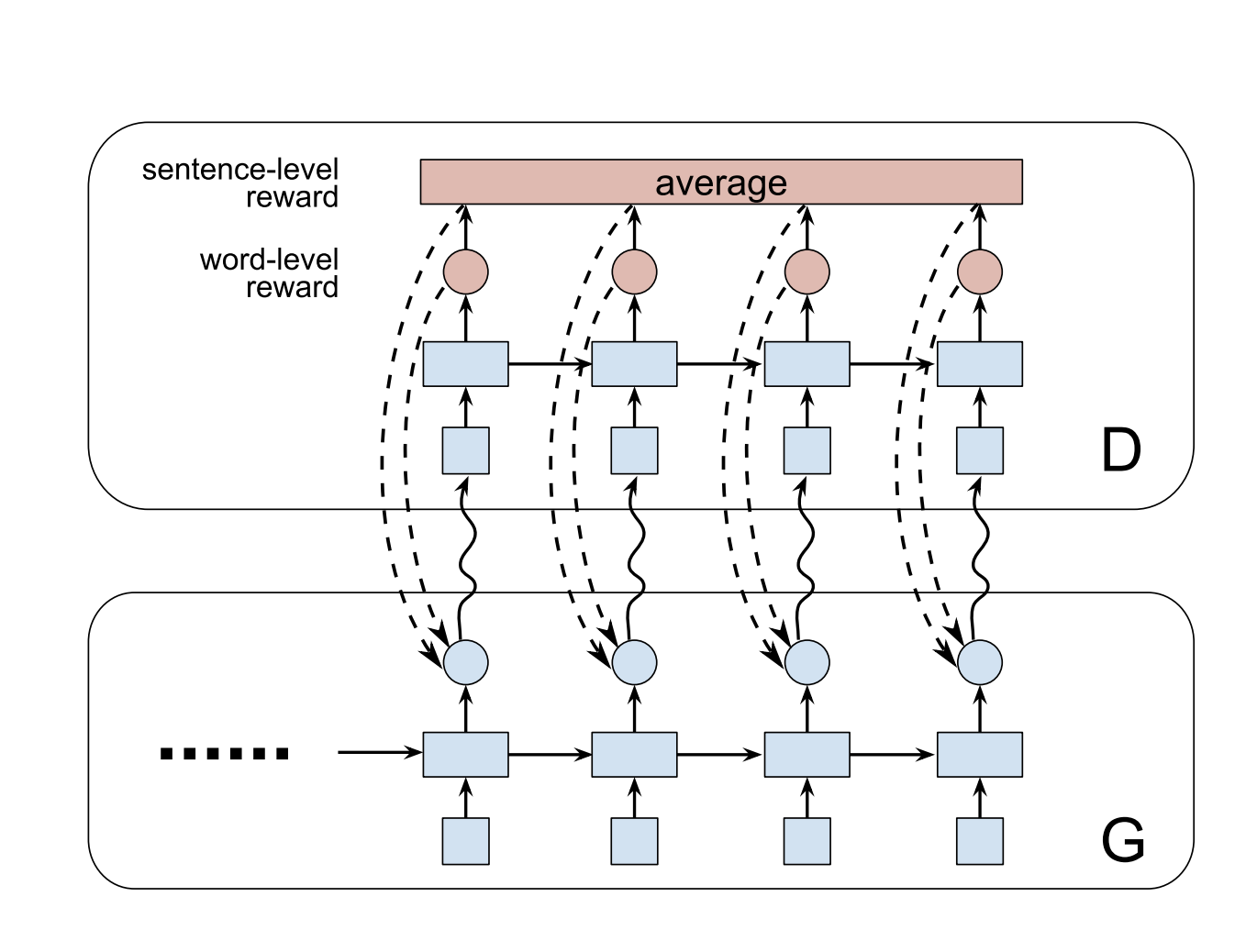}
    \caption{DP-GAN model , image from \cite{xu-etal-2018-diversity} }
\end{figure}

\textbf{Counterfactual Reasoning} is a psychology concept that describes human behaviour able to learn from previous experiences and create alternative solutions. It is a probabilistic answer to the question "what would have happened if ".  In the case of our work, counterfactual reasoning allows a bot to more accurately answer a question and participate to a discussion.  In the area of NLP, Deep Learning and reinforcement Learning, counterfactual reasoning is used for different purposes. Most commonly, it is used for data augmentation purposes \cite{kaushik2020learning}, \cite{zmigrod-etal-2019-counterfactual}. But also, in order to explore alternative policies that an agent could have been taken \cite{DBLP:journals/corr/abs-2004-14507}. But also, in the purpose of leveraging advantages of counterfactual reasoning for decision making in the reinforcement learning framework \cite{DBLP:journals/corr/abs-1811-06272}. Or another field of application of counterfactual reasoning is Learning representations as in \cite{johansson2018learning}. As extension of this work, applying a counterfactual inference system on the trained models should improve the diversity of the responses. \cite{DBLP:journals/corr/abs-2004-14507} implemented a CF framework on the REGS neural network. 

\bibliographystyle{apalike}
\bibliography{references}

\begin{thebibliography}{}

\bibitem[An et~al., 2021]{DBLP:journals/corr/abs-2104-03057}
An, C., Zhong, M., Chen, Y., Wang, D., Qiu, X., and Huang, X. (2021).
\newblock Enhancing scientific papers summarization with citation graph.
\newblock {\em CoRR}, abs/2104.03057.

\bibitem[Bahdanau et~al., 2016]{bahdanau2016neural}
Bahdanau, D., Cho, K., and Bengio, Y. (2016).
\newblock Neural machine translation by jointly learning to align and
  translate.

\bibitem[Brown et~al., 2020]{GPT}
Brown, T.~B., Mann, B., Ryder, N., Subbiah, M., Kaplan, J., Dhariwal, P.,
  Neelakantan, A., Shyam, P., Sastry, G., Askell, A., Agarwal, S.,
  Herbert-Voss, A., Krueger, G., Henighan, T., Child, R., Ramesh, A., Ziegler,
  D.~M., Wu, J., Winter, C., Hesse, C., Chen, M., Sigler, E., Litwin, M., Gray,
  S., Chess, B., Clark, J., Berner, C., McCandlish, S., Radford, A., Sutskever,
  I., and Amodei, D. (2020).
\newblock Language models are few-shot learners.

\bibitem[Buesing et~al., 2018]{DBLP:journals/corr/abs-1811-06272}
Buesing, L., Weber, T., Zwols, Y., Racani{\`{e}}re, S., Guez, A., Lespiau, J.,
  and Heess, N. (2018).
\newblock Woulda, coulda, shoulda: Counterfactually-guided policy search.
\newblock {\em CoRR}, abs/1811.06272.

\bibitem[Chen et~al., 2016]{NIPS2016_7c9d0b1f}
Chen, X., Duan, Y., Houthooft, R., Schulman, J., Sutskever, I., and Abbeel, P.
  (2016).
\newblock Infogan: Interpretable representation learning by information
  maximizing generative adversarial nets.
\newblock In Lee, D., Sugiyama, M., Luxburg, U., Guyon, I., and Garnett, R.,
  editors, {\em Advances in Neural Information Processing Systems}, volume~29.
  Curran Associates, Inc.

\bibitem[Chung et~al., 2014]{GRUPaper}
Chung, J., Gulcehre, C., Cho, K., and Bengio, Y. (2014).
\newblock Empirical evaluation of gated recurrent neural networks on sequence
  modeling.

\bibitem[Feng et~al., 2021]{feng2021survey}
Feng, S.~Y., Gangal, V., Wei, J., Chandar, S., Vosoughi, S., Mitamura, T., and
  Hovy, E. (2021).
\newblock A survey of data augmentation approaches for nlp.

\bibitem[Glover, 2016]{glover2016modeling}
Glover, J. (2016).
\newblock Modeling documents with generative adversarial networks.

\bibitem[Goodfellow et~al., 2014]{goodfellow2014generative}
Goodfellow, I.~J., Pouget-Abadie, J., Mirza, M., Xu, B., Warde-Farley, D.,
  Ozair, S., Courville, A., and Bengio, Y. (2014).
\newblock Generative adversarial networks.

\bibitem[Haidar and Rezagholizadeh, 2019]{haidar2019textkdgan}
Haidar, M.~A. and Rezagholizadeh, M. (2019).
\newblock Textkd-gan: Text generation using knowledgedistillation and
  generative adversarial networks.

\bibitem[Hochreiter and Schmidhuber, 1997]{LSTMPaper}
Hochreiter, S. and Schmidhuber, J. (1997).
\newblock {Long Short-Term Memory}.
\newblock {\em Neural Computation}, 9(8):1735--1780.

\bibitem[Johansson et~al., 2018]{johansson2018learning}
Johansson, F.~D., Shalit, U., and Sontag, D. (2018).
\newblock Learning representations for counterfactual inference.

\bibitem[Kaushik et~al., 2020]{kaushik2020learning}
Kaushik, D., Hovy, E., and Lipton, Z.~C. (2020).
\newblock Learning the difference that makes a difference with
  counterfactually-augmented data.

\bibitem[Labeau and Cohen, 2019]{labeau-cohen-2019-experimenting}
Labeau, M. and Cohen, S.~B. (2019).
\newblock Experimenting with power divergences for language modeling.
\newblock In {\em Proceedings of the 2019 Conference on Empirical Methods in
  Natural Language Processing and the 9th International Joint Conference on
  Natural Language Processing (EMNLP-IJCNLP)}, pages 4104--4114, Hong Kong,
  China. Association for Computational Linguistics.

\bibitem[Lamb et~al., 2016]{lamb2016professor}
Lamb, A., Goyal, A., Zhang, Y., Zhang, S., Courville, A., and Bengio, Y.
  (2016).
\newblock Professor forcing: A new algorithm for training recurrent networks.

\bibitem[Li et~al., 2015]{li2015hierarchical}
Li, J., Luong, M.-T., and Jurafsky, D. (2015).
\newblock A hierarchical neural autoencoder for paragraphs and documents.

\bibitem[Li et~al., 2017a]{Li2017AdversarialLF}
Li, J., Monroe, W., Shi, T., Jean, S., Ritter, A., and Jurafsky, D. (2017a).
\newblock Adversarial learning for neural dialogue generation.
\newblock {\em ArXiv}, abs/1701.06547.

\bibitem[Li et~al., 2017b]{DBLP:journals/corr/LiMSRJ17}
Li, J., Monroe, W., Shi, T., Ritter, A., and Jurafsky, D. (2017b).
\newblock Adversarial learning for neural dialogue generation.
\newblock {\em CoRR}, abs/1701.06547.

\bibitem[Li et~al., 2020]{LI2020103853}
Li, X., Grandvalet, Y., Davoine, F., Cheng, J., Cui, Y., Zhang, H., Belongie,
  S., Tsai, Y.-H., and Yang, M.-H. (2020).
\newblock Transfer learning in computer vision tasks: Remember where you come
  from.
\newblock {\em Image and Vision Computing}, 93:103853.

\bibitem[Li et~al., 2017c]{li2017dailydialog}
Li, Y., Su, H., Shen, X., Li, W., Cao, Z., and Niu, S. (2017c).
\newblock Dailydialog: A manually labelled multi-turn dialogue dataset.

\bibitem[Luan et~al., 2016]{luan2016lstm}
Luan, Y., Ji, Y., and Ostendorf, M. (2016).
\newblock Lstm based conversation models.

\bibitem[Luketina et~al., 2019]{luketina2019survey}
Luketina, J., Nardelli, N., Farquhar, G., Foerster, J., Andreas, J.,
  Grefenstette, E., Whiteson, S., and Rocktäschel, T. (2019).
\newblock A survey of reinforcement learning informed by natural language.

\bibitem[Miao and Blunsom, 2016]{miao-blunsom-2016-language}
Miao, Y. and Blunsom, P. (2016).
\newblock Language as a latent variable: Discrete generative models for
  sentence compression.
\newblock In {\em Proceedings of the 2016 Conference on Empirical Methods in
  Natural Language Processing}, pages 319--328, Austin, Texas. Association for
  Computational Linguistics.

\bibitem[Radford et~al., 2016]{radford2016unsupervised}
Radford, A., Metz, L., and Chintala, S. (2016).
\newblock Unsupervised representation learning with deep convolutional
  generative adversarial networks.

\bibitem[Raffel et~al., 2020]{raffel2020exploring}
Raffel, C., Shazeer, N., Roberts, A., Lee, K., Narang, S., Matena, M., Zhou,
  Y., Li, W., and Liu, P.~J. (2020).
\newblock Exploring the limits of transfer learning with a unified text-to-text
  transformer.

\bibitem[Rajeswar et~al., 2017]{rajeswar2017adversarial}
Rajeswar, S., Subramanian, S., Dutil, F., Pal, C., and Courville, A. (2017).
\newblock Adversarial generation of natural language.

\bibitem[Ramamurthy et~al., 2020]{ramamurthy2020nlpgym}
Ramamurthy, R., Sifa, R., and Bauckhage, C. (2020).
\newblock Nlpgym -- a toolkit for evaluating rl agents on natural language
  processing tasks.

\bibitem[Ranzato et~al., 2016]{ranzato2016sequence}
Ranzato, M., Chopra, S., Auli, M., and Zaremba, W. (2016).
\newblock Sequence level training with recurrent neural networks.

\bibitem[Ritter et~al., 2011]{ritter-etal-2011-data}
Ritter, A., Cherry, C., and Dolan, W.~B. (2011).
\newblock Data-driven response generation in social media.
\newblock In {\em Proceedings of the 2011 Conference on Empirical Methods in
  Natural Language Processing}, pages 583--593, Edinburgh, Scotland, UK.
  Association for Computational Linguistics.

\bibitem[Ruder et~al., 2019]{ruder-etal-2019-transfer}
Ruder, S., Peters, M.~E., Swayamdipta, S., and Wolf, T. (2019).
\newblock Transfer learning in natural language processing.
\newblock In {\em Proceedings of the 2019 Conference of the North {A}merican
  Chapter of the Association for Computational Linguistics: Tutorials}, pages
  15--18, Minneapolis, Minnesota. Association for Computational Linguistics.

\bibitem[Rumelhart and McClelland, 1987]{RNNReport}
Rumelhart, D.~E. and McClelland, J.~L. (1987).
\newblock {\em Learning Internal Representations by Error Propagation}, pages
  318--362.

\bibitem[Schmidt, 2019]{Schmidt19}
Schmidt, F. (2019).
\newblock Generalization in generation: A closer look at exposure bias.
\newblock In {\em NGT@EMNLP-IJCNLP}, pages 157--167.

\bibitem[Schuster and Paliwal, 1997]{BiDir}
Schuster, M. and Paliwal, K. (1997).
\newblock Bidirectional recurrent neural networks.
\newblock {\em IEEE Transactions on Signal Processing}, 45(11):2673--2681.

\bibitem[Serban et~al., 2016]{serban2016generative}
Serban, I.~V., Lowe, R., Charlin, L., and Pineau, J. (2016).
\newblock Generative deep neural networks for dialogue: A short review.

\bibitem[Shen et~al., 2018]{shen2018improving}
Shen, X., Su, H., Niu, S., and Demberg, V. (2018).
\newblock Improving variational encoder-decoders in dialogue generation.

\bibitem[Sutskever et~al., 2014]{sutskever2014sequence}
Sutskever, I., Vinyals, O., and Le, Q.~V. (2014).
\newblock Sequence to sequence learning with neural networks.

\bibitem[Vaswani et~al., 2017]{vaswani2017attention}
Vaswani, A., Shazeer, N., Parmar, N., Uszkoreit, J., Jones, L., Gomez, A.~N.,
  Kaiser, L., and Polosukhin, I. (2017).
\newblock Attention is all you need.

\bibitem[Vinyals and Le, 2015]{vinyals2015neural}
Vinyals, O. and Le, Q. (2015).
\newblock A neural conversational model.

\bibitem[Wen et~al., 2017]{pmlr-v70-wen17a}
Wen, T.-H., Miao, Y., Blunsom, P., and Young, S. (2017).
\newblock Latent intention dialogue models.
\newblock In Precup, D. and Teh, Y.~W., editors, {\em Proceedings of the 34th
  International Conference on Machine Learning}, volume~70 of {\em Proceedings
  of Machine Learning Research}, pages 3732--3741. PMLR.

\bibitem[Williams, 1992]{Williams92simplestatistical}
Williams, R.~J. (1992).
\newblock Simple statistical gradient-following algorithms for connectionist
  reinforcement learning.
\newblock In {\em Machine Learning}, pages 229--256.

\bibitem[Xu et~al., 2018]{xu-etal-2018-diversity}
Xu, J., Ren, X., Lin, J., and Sun, X. (2018).
\newblock Diversity-promoting {GAN}: A cross-entropy based generative
  adversarial network for diversified text generation.
\newblock In {\em Proceedings of the 2018 Conference on Empirical Methods in
  Natural Language Processing}, pages 3940--3949, Brussels, Belgium.
  Association for Computational Linguistics.

\bibitem[Yu et~al., 2017]{yu2017seqgan}
Yu, L., Zhang, W., Wang, J., and Yu, Y. (2017).
\newblock Seqgan: Sequence generative adversarial nets with policy gradient.

\bibitem[Zhu et~al., 2020]{DBLP:journals/corr/abs-2004-14507}
Zhu, Q., Zhang, W., Liu, T., and Wang, W.~Y. (2020).
\newblock Counterfactual off-policy training for neural response generation.
\newblock {\em CoRR}, abs/2004.14507.

\bibitem[Zmigrod et~al., 2019]{zmigrod-etal-2019-counterfactual}
Zmigrod, R., Mielke, S.~J., Wallach, H., and Cotterell, R. (2019).
\newblock Counterfactual data augmentation for mitigating gender stereotypes in
  languages with rich morphology.
\newblock In {\em Proceedings of the 57th Annual Meeting of the Association for
  Computational Linguistics}, pages 1651--1661, Florence, Italy. Association
  for Computational Linguistics.

\end{thebibliography}

\end{document}